\documentclass[10pt,twocolumn,letterpaper]{article}

\usepackage{iccv}
\usepackage{times}
\usepackage{epsfig}
\usepackage{graphicx}
\usepackage{amsmath}
\usepackage{amssymb}
\usepackage{enumitem}
\usepackage{algorithmic}
\usepackage{algorithm}

\usepackage[pagebackref=true,breaklinks=true,letterpaper=true,colorlinks,bookmarks=false]{hyperref}

\iccvfinalcopy 

\def\iccvPaperID{1041} 

\def\bg{{\tt bg}}
\def\fg{{\tt f\!g}}

\def\bx{{\mathbf{x}}}
\def\bc{{\mathbf{c}}}
\def\bb{{\mathbf{b}}}

\def\bo{{\mathbf{0}}}
\def\mNB{{\mathcal{N_B}}}

\def\SigCB{{\Sigma^{B}_{\mathbf C}}}
\def\SigXB{{\Sigma^{B}_{\mathbf S}}}
\def\SigCF{{\Sigma^{F}_{\mathbf C}}}
\def\SigXF{{\Sigma^{F}_{\mathbf S}}}

\newcommand{\argmax}{\operatornamewithlimits{argmax}}
\ificcvfinal\fi
\begin{document}

\title{Coherent Motion Segmentation in Moving Camera Videos\\
    using Optical Flow Orientations}

\author{Manjunath Narayana\\
{\tt\small narayana@cs.umass.edu}
\and
Allen Hanson\\
{\tt\small hanson@cs.umass.edu}\\
University of Massachusetts, Amherst\\
\and
Erik Learned-Miller\\
{\tt\small elm@cs.umass.edu}
}
\maketitle

\begin{abstract}
   In moving camera videos, motion segmentation is commonly 
   performed using the image plane motion of pixels, or optical flow. 
   However, objects that are at different depths from the camera can exhibit
   different optical flows even if they share the same real-world motion.
   This can cause a depth-dependent segmentation of the scene.
   Our goal is to develop a segmentation algorithm that clusters  
   pixels that have similar real-world motion irrespective of their depth
   in the scene.
   Our solution uses optical flow orientations instead of the complete 
   vectors and exploits the well-known property that under 
   camera translation, optical flow orientations are 
   independent of object depth.
   We introduce a probabilistic model that automatically estimates 
   the number of observed independent motions and results in a 
   labeling that is consistent with real-world motion in the scene.
   The result of our system is that static objects are correctly 
   identified as one segment, even if they are at different depths.
   Color features and information from previous 
   frames in the video sequence are used to correct occasional errors due to 
   the orientation-based segmentation.
   We present results on more than thirty videos from different benchmarks.
   The system is particularly robust on complex background scenes containing 
   objects at significantly different depths.
\end{abstract}

\begin{figure}[t]
\begin{center}
   \includegraphics[width=0.7\linewidth]{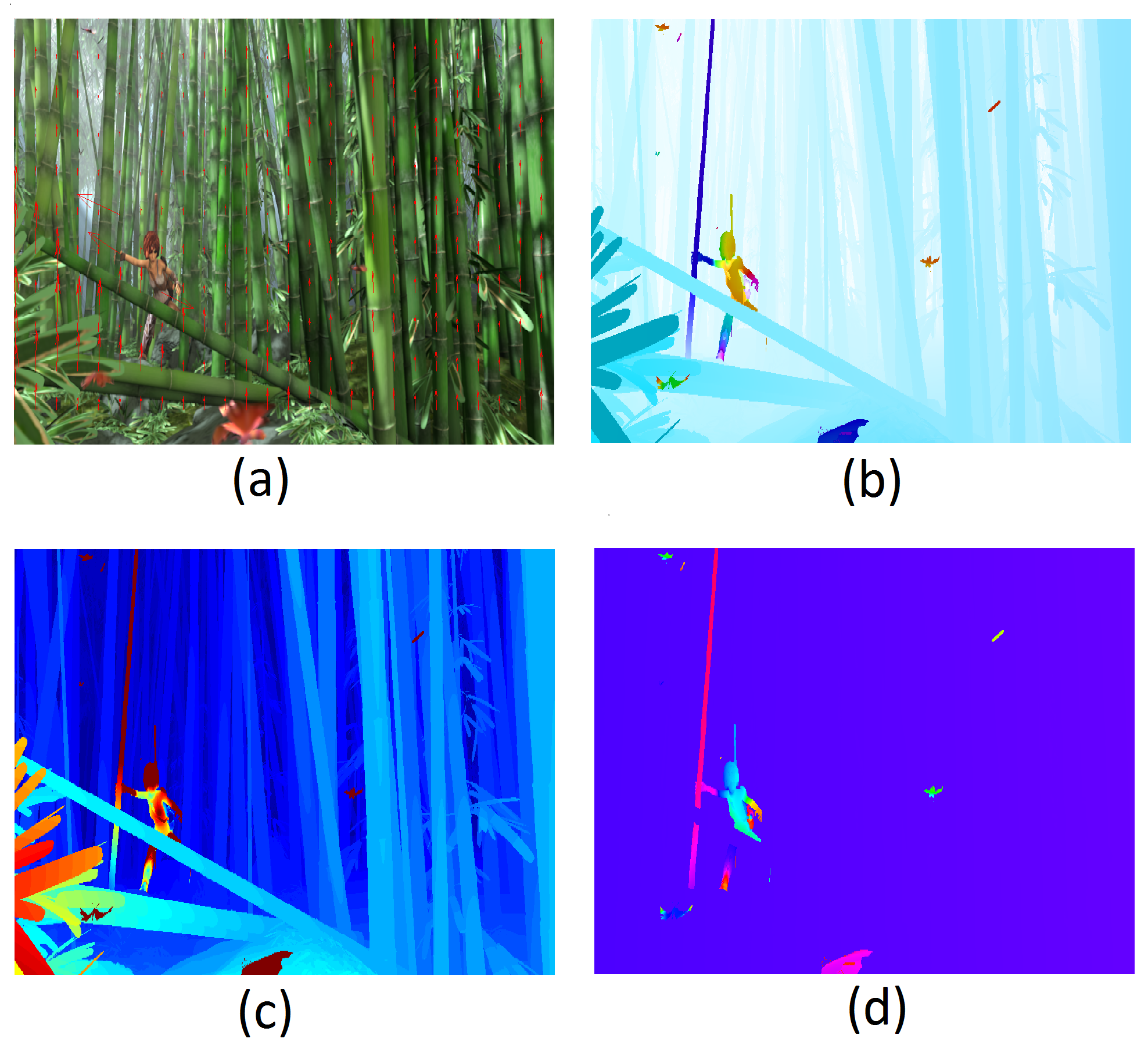}
\end{center}
   \caption{(a) A forest scene with a moving person (from the 
           Sintel~\cite{Butler12} data set). 
       The person is holding on to a bamboo tree, which
       moves with the person. There are also a few leaves falling in the
       scene.
       (b) Visualization of the ground truth optical flow vectors 
       (using code from ~\cite{Sun10}).
        (c) Magnitudes of the optical flow vectors.
        (d) Orientation of the optical flow vectors. 
        The optical flow vectors and magnitudes on the trees depend on the 
        distance of the trees from the camera.
        The orientations are not depth-dependent and can much more reliably
        predict that all the trees are part of the coherently moving 
        background entity.
       }
\label{fig:angleFieldEx}
\end{figure}

\section{Introduction}

   Motion segmentation in stationary camera videos is relatively straightforward 
   and a pixelwise background model can accurately classify pixels  
   as background or foreground. 
   The pixelwise models may be built using a variety of techniques 
   such as the mixture of Gaussians 
   model~\cite{Stauffer99}, kernel density estimation~\cite{Elgammal00},
   and joint domain-range modeling~\cite{Sheikh05, Narayana12BMVC}. 
   While background segmentation for stationary cameras can be estimated 
   accurately, separating the non-moving objects from
   moving ones when the camera is moving is significantly more
   challenging.
   Since the camera's motion causes most image pixels to move, pixelwise
   models are no longer adequate. 
   A common theme in moving camera motion segmentation is to use
   image plane motion (\emph{optical flow}) or trajectories 
   as a surrogate 
   for real-world object motion.
   Image plane motion can be used directly as a cue for 
   clustering~\cite{Shi98, Sheikh09, Brox10, Kwak11, Elqursh12, Ochs12} 
   or to compensate for the camera motion so that the 
   pixelwise model from the previous frame can be adjusted in order 
   to remain accurate~\cite{Irani94, Hayman03, Ren03}.

   The major drawback of using optical flow is that an object's projected
   motion on the image plane depends on the object's distance from the camera.
   Objects that have the same real-world motion can have different
   optical flows depending on their depth. 
   This can cause a clustering algorithm to label 
   two background objects at different depths as two separate objects
   although they both have zero motion in the real-world.
   While this labeling is semantically consistent because the two 
   segments are likely to correspond to different
   objects in the world, such over-segmentation of the scene is undesirable
   for the purpose of detecting independently moving
   objects in the scene.
   For example, in Figure~\ref{fig:angleFieldEx}, the optical flow vectors
   separate the forest background into many smaller tree segments.
   Post-processing is required to merge smaller segments into one background
   cluster.
   Existing algorithms merge segments 
   based on their color, motion, and edge energy.
   If the number of distinct background layers is known, mixture modeling of the 
   background motion is another solution.
   
   An ideal solution would not require the use of post-processing of segments
   or prior knowledge about the scene. Our goal is to segment the scene into
   coherent regions based on the real-world motion of the objects in it. 
   This can be challenging since the information about 3-D motion in the 
   scene is only available in the form of the optical flow field.
   Our solution is based on the well-known property that for translational 
   camera motion, while optical flow 
   magnitudes and vectors depend on the depth of the object in the scene,
   the orientations of the optical flow vectors do not.
   Figure~\ref{fig:angleFieldEx} is an example that shows that the 
   optical flow orientations are reliable 
   indicators of real-world motion, much more so than the flow vectors or 
   magnitudes. 

   Assuming only translational motions in the scene, 
   given the motion parameters of
   the objects and knowledge about which pixels belong to each 
   object, it is straightforward to predict the orientations 
   at each pixel exactly. 
   Figure~\ref{fig:FOF} shows some examples of such predicted  
   orientation \emph{fields} for different motion parameter values.
   Our problem is the converse:
   Given the observed optical flow orientations at each pixel, estimate 
   the motion parameters and pixel labels. 
   We solve the problem by starting with a 
   ``library'' of predicted orientation \emph{fields} which cover a large 
   space of possible translations and then use a probabilistic 
   model to estimate which of these 
   predicted orientation fields are actually being observed in the current 
   image.
   Since multiple motions (one camera motion and possibly other independent
   object motions) are possible, we use a mixture model to determine which 
   motions are present in the scene and which pixels belong to each motion.
   Finally, we favor explanations with fewer 3-D motions.
   A similar system involving optical flow magnitudes
   is much more complicated because in addition to estimating the motion
   parameters, it would be required to determine the object depth at each 
   pixel.
   
   Performing clustering when the number of foreground objects is unknown can be 
   challenging. Techniques such as K-means or expectation maximization 
   (EM) require knowing the number of clusters before-hand.
   We avoid this problem by instead using a non-parametric Dirichlet 
   process-like mixture model where the number of components is 
   determined automatically.
   Our system is capable of segmenting background objects at different depths
   into one segment and identifying the various regions that
   correspond to coherently moving foreground segments.
   
   Although the optical flow orientations are effective in many scenarios,
   they are not always reliable. Our algorithm 
   is prone to failure when the assumption of pure translation is 
   violated.
   Also, a foreground object that moves in a direction consistent with the 
   flow orientations due to the camera's motion will go undetected until
   it changes its motion direction.
   These occasional errors are handled in our system through the use of
   a pixelwise color appearance model.

   Earlier approaches to motion segmentation with a moving camera relied
   on motion compensation~\cite{Irani94, 
       Hayman03, Ren03} after estimating the camera's motion
   as a 2-D affine transformation or a homography.
   These techniques work well when the background can be approximated
   as a planar surface. 
   More recent techniques have performed segmentation by clustering
   the trajectory information from multiple frames
   ~\cite{Sheikh09, Brox10, Elqursh12, Ochs12}.
   Sheikh~\etal~\cite{Sheikh09} use a factorization method to find the 
   bases for the background trajectories and label outlier trajectories
   as foreground.
   However, they assume an orthographic camera model. 
   Brox and Malik~\cite{Brox10} segment trajectories by computing the
   pairwise distances between all trajectories and finding a low-dimensional
   embedding using spectral clustering. Their method is not online and 
   works on the video by considering all or a subset of frames at once.
   
   Ochs and Brox~\cite{Ochs12} improved the spectral clustering by using 
   higher order interactions that consider triplets of trajectories.
   Elqursh and Elgammal~\cite{Elqursh12} proposed an online extension of 
   spectral clustering by considering trajectories from $5$ frames at a time.
   Because they rely on distance between optical flow vectors, these 
   spectral methods are not guaranteed to group all the 
   background pixels into one cluster. 
   To obtain the complete background as one segment, a 
   post-processing merging step is required where segments with similar motions 
   are merged~\cite{Brox10,Ochs12}. 
   The merging step assumes an affine motion model and hence may
   not work for complex backgrounds, as we show in Section~\ref{sec:SegCompare}. 
   Elqursh and Elgammal learn a mixture of $5$ Gaussians in the embedded 
   space to represent the trajectories. 
   Any trajectory that is not well explained by
   the mixture of Gaussians model is assumed to be a foreground trajectory.
   The parametric Gaussian mixtures model requires the number of mixtures, 
   which can vary from scene to scene.
   
   A significant improvement over the simple appearance and tracking model
   in the above papers was proposed by Kwak~\etal~\cite{Kwak11}. They use a Bayesian 
   filtering framework that combines block-based color appearance models
   with separate motion models for the background and foreground to 
   estimate the labels at each pixel. However, they use a special 
   initialization procedure in the first frame for segmenting the foreground 
   objects.
   Their initialization procedure and the earlier trajectory-based methods 
   use image plane motion. As described earlier, this cue is prone to 
   causing errors.
   
   In comparison to the above methods, we use motion information only from
   two frames at a time and do not require the use of trajectory 
   information from multiple frames.
   In contrast to Kwak~\etal, our system is completely online, with no
   special initialization step for the first frame.
   Due to automatic determination of the number of observed motions, our system 
   is able to detect objects that are at rest initially and which
   begin to move during the video sequence.
   
   Object tracking in a moving camera video is another theme in recent work. 
   Chockalingam~\etal~\cite{Chockalingam09} learn a fragmented model of
   the scene by breaking the image into smaller fragments which are 
   then assigned foreground/background labels and tracked in subsequent 
   frames. 
   Tsai~\etal~\cite{Tsai10} achieve tracking by using a spatio-temporal 
   Markov random field (MRF) and introducing pairwise potentials that 
   represent appearance and motion similarity between neighboring
   pixels. These tracking systems require an initial human-labeled 
   foreground object while our goal is to build a foreground-background
   segmentation algorithm without any human intervention. 
   Lee~\etal~\cite{Lee11} detect object-like
   segments called \emph{key-segments} in the image, hypothesize which 
   segments are more likely to be foreground objects, and finally use a 
   spatio-temporal graph to perform segmentation. Although they avoid the
   requirement of hand-labeling the object of interest, their method is
   suited for offline processing of videos because the initial 
   key-segments generation phase requires the processing of all frames of the
   video. Our goal is to process a video frame-by-frame as they appear in 
   the video stream.
  
   Earlier background segmentation methods report results only on
   3 or 4 out of 26 videos from the Hopkins segmentation data 
   set~\cite{Brox10}.
   In addition to all 26 videos from this set, we also include results 
   from the SegTrack motion segmentation data set~\cite{Tsai10}. 
   Although good segmentation results are achieved on 
   these data sets, these videos have few cases of depth disparity in the 
   background. 
   Consequently, results from other videos with complex backgrounds that 
   can involve many many depth layers, such as in a forest scene, are also 
   presented.
   To the best of our knowledge, this is the first work to report moving 
   background segmentation results on such a large number of videos spanning
   different scenarios.
   The results show the efficacy of the algorithm and its applicability to a 
   wide range of videos. Despite the assumption of translational camera 
   motion, the algorithm is capable of handling many scenarios as exhibited 
   in the data set.
   


\section{Segmentation using optical flow orientations}\label{sec:FOF}

Given a camera's translation $t=(t_x, t_y, t_z)$, 
the resulting optical flows $v_x$ and $v_y$ in the $x$ and $y$ 
image dimensions are given by:
\begin{equation}\label{eq:vuFlow}
v_x = \frac{t_z\!\times\! x - t_x\!\times\! f}{Z}
\quad\text{and}\quad
v_y = \frac{t_z\!\times\! y - t_y\!\times\! f}{Z},
\end{equation}
where $(x, y)$ represents a pixel location in the image, $Z$ is the
real-world depth of the observed point 
and $f$ is the camera's focal length~\cite{Irani94CVPR}.

\begin{figure}[t]
\begin{center}
   \includegraphics[width=0.7\linewidth]{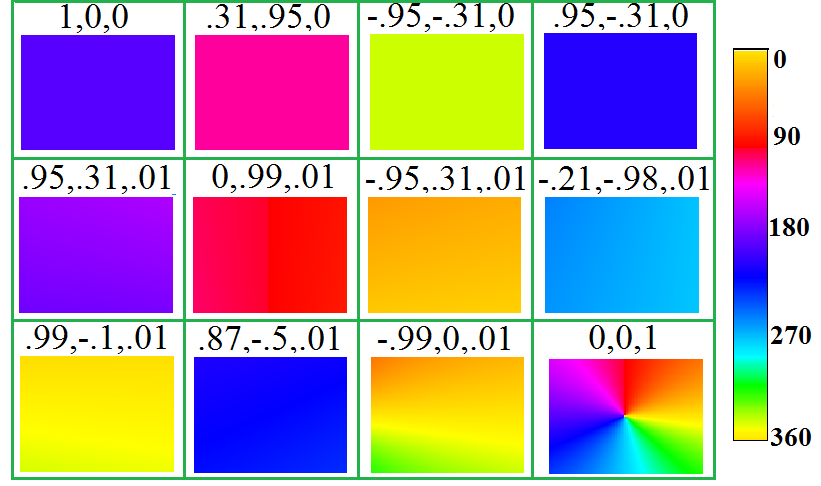}
\end{center}
   \caption{A sample set from the orientation fields that are used in 
       our graphical model. 
       Above each field are the motion parameters ($t_x,t_y,t_z$) that cause it.
       The colorbar on the right
       shows the mapping from color values to corresponding angles 
       in degrees. 
       }
\label{fig:FOF}
\end{figure}

The optical flow orientations,
\begin{equation}\label{eq:vuAngle}
F(t,x,y)   = 
          \arctan(t_z\!\times\! y-t_y\!\times\! f, t_z\!\times\! x-t_x\!\times\! f),
\end{equation}
are thus independent of the depth $Z$ of the points.
Here, $\arctan(y,x)$ returns the arctangent of $(y/x)$ with a range
$(-\pi,\pi]$.
Figure~\ref{fig:FOF} shows the optical flow orientations for
a few different camera motions. It may be noted that the 
orientations are not always constant throughout the entire 
image. We call the 2-D matrix of optical flow orientations
at each pixel the \emph{flow orientation field} (FOF).


\begin{figure}[t]
\begin{center}
   \includegraphics[width=0.6\linewidth]{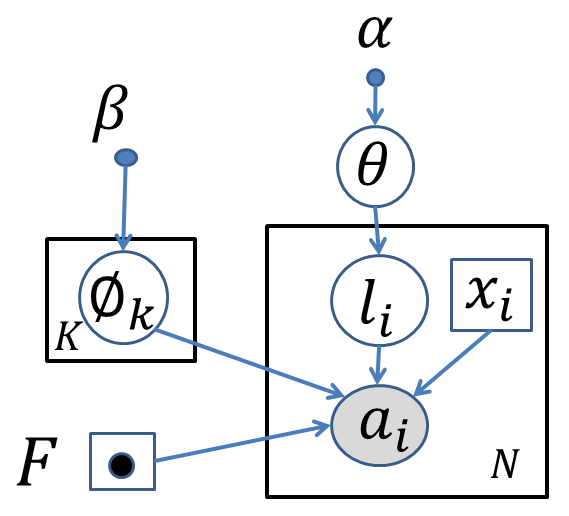}
\end{center}
   \caption{A mixture model for segmentation based on optical
       flow orientations. Notation: Variables inside
   circles are random variables and variables inside squares
   are deterministic. The dark colored dot represents a 
   deterministic function, the shaded circle represents an
   observed variable and small shaded circles represent
   hyperparameters.}
\label{fig:topicModel}
\end{figure}

In the probabilistic model given in Figure~\ref{fig:topicModel},
the orientation values returned by an optical flow estimation algorithm
~\cite{Sun10}
are the observed variables and the labels for each pixel are latent. 
At pixel number $i$, whose location is given by $\bx_i=(x_i,y_i)$, 
we have an observed optical flow orientation $a_i$ 
and a label $l_i$ that represents which segment the pixel belongs to.
Each segment $k$ is associated with a motion parameter tuple 
$\Phi_k = (t^k_x, t^k_y, t^k_z)$ representing the translation
along $x$, $y$, and $z$ directions respectively.
The continuous velocity space is discretized into 
a finite number of possible motions:
$46$ values for translation $(t_x,t_y,t_z)$ are sampled
from a unit hemisphere in front of an observer.
$\Phi_k$ can hence take one of $46$ values and the resulting
FOFs due to these motion values form a ``library'' that is
used to explain the observed data. Figure~\ref{fig:FOF} shows a
few of the library FOFs; a complete listing of the FOFs used is 
provided in the supplementary submission.
$\phi_k$ is used to denote the values that the variables $\Phi_k$ take.
For a given motion parameter tuple $t$, denote the resulting flow 
orientation field at pixel location $\bx$ to be $F(t,\bx)$, which
is computed using Equation~\ref{eq:vuAngle}.

The graphical model is then defined by the following generative process:
\begin{equation}\label{eq:GmEqns}
\begin{split}
P(\theta|\alpha) &= \text{Dir}(\theta|\alpha);\\
P(\Phi_k|\beta) &= \text{Uniform}(\beta);\\
P(l_i|\theta) &= \prod_{k=1}^K\theta_k^{[l_i=k]};\\
P(a_i|\Phi=\phi,l_i=k,F,\bx_i) &= P(a_i|\Phi_k=\phi_k,F(\phi_k,\bx_i))\\
        &= G(a_i;F(\phi_k,\bx_i),\sigma^2_k),
\end{split}
\end{equation}
where $[\cdot]$ represents an indicator function, $\text{Dir}$ is a Dirichlet 
distribution, and $G(.;\mu,\sigma^2)$ is a Gaussian  
with mean $\mu$ and variance $\sigma^2$.
The last equation means that given the label $l_i=k$ for a pixel at 
location $\bx_i$ and motion parameter $\Phi_k=\phi_k$, 
the observed orientation $a_i$ is a Gaussian random variable 
whose mean is $F(\phi_k,\bx_i)$. The variance for the Gaussian is the
observed variance from $F(\phi_k,\bx'_i)$ at all pixel locations $\bx'$ 
that were labeled $k$ in the previous iteration.
If no pixels were labeled $k$ in the 
previous iteration, a variance value of $(a_\bx-F_\bx(\phi_k))^2$ is used.

We note that the above model is similar to a Dirichlet process
mixture model with the exception that we sample $\Phi_k$ from a finite 
set of parameters. 
For sampling, a Gibbs sampling algorithm that introduces
auxiliary parameters at each iteration is used, similar to  
algorithm 8 from Neal~\cite{Neal00}
(detailed in the supplementary submission).
The algorithm adds additional auxiliary $\Phi$ parameters at each 
iteration and retains the auxiliary parameters that explain any 
observed data.
We begin with $K=1$
component and add one new auxiliary component at each 
iteration. 
The model hence adds components as required to explain the data.
\subsection{Choosing $\alpha$}
The concentration parameter $\alpha$ determines the propensity of the
system to add new components.
In the absence of suitable training data to learn the concentration 
parameter, the Gibbs sampler is run with different values for
$\alpha_j\in\{ .0001, .01, 10\}$
and, from the resulting segmented images, the segmented image that best
agrees with the other segmented images is chosen.
From each candidate $\alpha_j$, the segmented result is obtained  
and an image $b_j(\bx)$, which has a value $1$
at locations that correspond to the largest segment and $0$ at all other
locations, is created.
The sum of these $b_j$ images is then computed: 
   $b_{\text{sum}}(\bx)=\sum_{j=1}^{n_{\alpha}} b_j(\bx)$,
where $n_{\alpha}$ is the number of different $\alpha$'s being considered. 
Similarly, $f_j$ and $f_\text{{sum}}$ images are computed, where $f_j= 1-b_j$. 
The best $\alpha$ corresponds to $\hat{j} = \argmax_j \sum_\bx \{b_\text{{sum}}(\bx)\times b_j(\bx)\} +\{f_\text{{sum}}(\bx)\times f_j(\bx) \}$.
Intuitively, $b_\text{{sum}}$ and $f_\text{{sum}}$ are the 
pixelwise sum of the votes for 
the background and foreground from all candidate $\alpha$'s. 
The best $\alpha$ is the one that best agrees with this voting.

\subsection{Gradient descent for largest component}
Improvements can be made to the results by finding a better fit for the 
largest segment's motion than provided by the relatively coarse initial
sampling of library motion parameters. 
To achieve this, after $\frac{n}{2}$ iterations, at
each iteration, we follow the Gibbs sampling step with a gradient descent step.
With the motion parameters corresponding to the largest segment as the
starting point, gradient descent is used to find the motion parameters
that result in an FOF with minimum average L1 distance to the observed 
orientations. Only the pixels that are currently assigned to the largest segment
are used in computing the L1 distance. The resulting minimum motion parameter
tuple is added as an additional motion parameter to the set 
of library motions.
This process helps in the proper segmentation of observed background 
orientation patterns that are not well explained by any of the initial 
set of motions.

\subsection{Handling pixels with near-zero motion}
One of the implications of using the orientations is that the orientation
is not defined for pixels that do not move. The orientation values at 
these pixels can be very noisy. To account for this possibility, pixels
that have optical flow component magnitudes less than a threshold 
$T_f$ (typically $0.5$) in both $x$ and $y$ directions are 
marked as ``zero-motion'' pixels. 
They are accounted for by a ``zero-motion'' FOF and Gibbs sampling
is not performed for these pixels.

\section{Segmentation comparisons}\label{sec:SegCompare}
\begin{figure*}[t]
\begin{center}
    \includegraphics[width=0.75\linewidth]{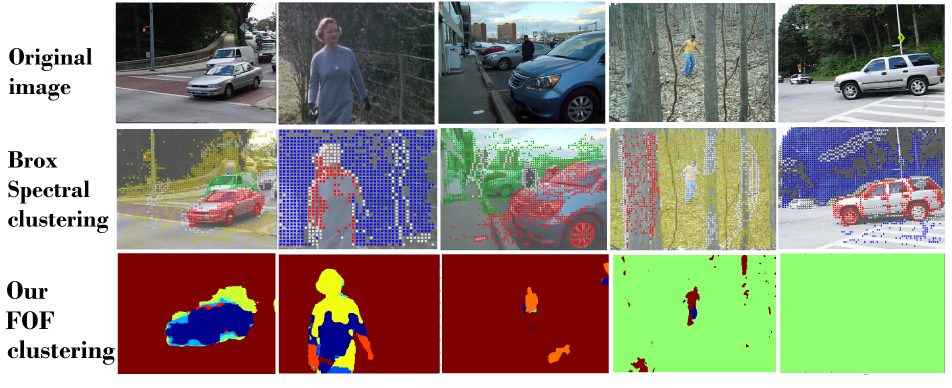}
\end{center}
   \caption{Comparison of segmentation algorithms. The rows correspond to
       the original images, spectral clustering~\cite{Ochs12}, and our
       FOF segmentation.
       The tracked keypoints used in spectral clustering are shown as squares 
       with their colors representing the cluster memberships. 
       Despite the use of a post-processing merge step in 
       the implementation, in many images, spectral clustering is not 
       certain about some background keypoints (white squares) and in 
       cases with large depth disparity, the background is broken into 
       smaller sections.
       Our method avoids these errors and also results in a dense labeling of
       the image.
       The last column is an example of our method failing because the
       car happens to move consistently with the FOF due to camera motion.
       More comparisons are provided in the supplementary submission.}
\label{fig:segCompare}
\end{figure*}

The proposed FOF segmentation is compared to existing motion segmentation methods. 
Spectral clustering of trajectory 
information~\cite{Brox10,Elqursh12,Ochs12} 
has been shown to be useful for motion segmentation.
The implementation provided by Ochs and Brox~\cite{Ochs12} that 
returns spectral clustering of tracked keypoints is used. 
Their algorithm is designed to work on
trajectories from multiple frames.
The number of frames is set to 3 for trajectory tracking 
(the minimum that their implementation requires). 
Further, their method uses a merging step that joins segments
that have similar motion parameters.
Note that FOF segmentation uses only flow information from two consecutive 
frames and performs no post-processing to merge segments.
Figure \ref{fig:segCompare} shows the segmentations for some example frames.
FOF segmentation, despite only using information from two frames and no merging
procedure, successfully segments the background in most examples.
Images that have large depth disparity show the clear advantage of our 
method (columns 3 and 4). Here spectral clustering with a subsequent 
merge step fails and the background is over-segmented depending on depth. 
The FOF-based clustering is successful in 
identifying the background objects as one segment. 

\section{Appearance modeling}\label{sec:workflow}
The described FOF-based mixture model returns 
the number of mixture components,
the maximum a posteriori component assignments for each pixel, and
the probabilities of each pixel belonging to each component.
In order to classify each pixel as background or foreground, 
the component with the largest number of pixels is considered as 
the background component.

In addition to using the FOF-based segmentation, we maintain a color 
appearance model for the background and foreground at each pixel~\cite{Sheikh09}.
A history of pixel data samples from the previous frames is
maintained and after classification of pixels in each new frame,
new data samples are added to the history. To account for motion,
the maintained history at each pixel is motion compensated and moved to 
a new location as predicted by the optical flow in the current frame.
Kernel density estimation (KDE) is used with the data samples to obtain the 
color likelihoods for the background and foreground processes.
To allow for spatial uncertainty in a pixel's location, we use data 
samples not only from the same (motion compensated) pixel location 
but from a small spatial neighborhood around that location.
This use of information from a pixel's neighborhood has been shown to
improve the accuracy of background modeling systems~\cite{Sheikh05,Narayana12BMVC}.
First let us consider single frame history.
Let $\bc$ represent the color vector $(r,g,b)$ of red, green, and blue intensities
respectively.
Let $\bb^{t-1}_{\bx}$ be the observed background color at 
pixel location $\bx$ in the previous frame.
Using a Gaussian kernel with covariance $\SigCB$ in the color dimensions,
our KDE background likelihood for the color vector $\bc$ in the video frame 
numbered $t$ is given by
\begin{equation}\label{eq:KDE_P_BG} 
\begin{split}
P^t_\bx(\bc|\bg;\SigCB,\SigXB) = \frac{1}{Z}
       \sum_{\Delta\in \mNB} &(G(\bc-\bb^{t-1}_{\bx+\Delta}; \bo, \SigCB)\\
                               &\times G(\Delta; \bo, \SigXB)).
\end{split}
\end{equation}
$\Delta$ is a spatial
displacement that defines a spatial neighborhood $\mNB$ around the pixel 
location $\bx$ at which the likelihood is being computed. 
$G(\cdot; \bo,\SigXB)$ is a zero-mean multivariate Gaussian with 
covariance $\SigXB$. $B$ indicates that the covariance 
is for the background model and $\mathbf{S}$ denotes the spatial dimension.
The covariance matrix $\SigXB$ controls the amount 
of spatial influence from neighboring pixels. The covariance matrix
$\SigCB$ controls the amount of variation allowed in the color values of
the background pixels.
The normalization constant $Z$ is 
\begin{equation}\label{eq:Z_norm} 
Z = \sum_{\Delta\in \mNB} G(\Delta; \bo, \SigXB).
\end{equation}

Considering background data samples not just from the previous
frame, but from the previous $T$ frames, and allowing probabilistic 
contribution from the previous 
frames' pixels, we have
\begin{equation}\label{eq:DF_KDE_bg} 
\begin{split}
P^t_\bx(\bc|\bg;\!\Sigma^B)\!\! =&\!\!\\
        \frac{1}{K_\bg}\!\!\sum_{i\in 1:T}\! \sum_{\Delta\in \mNB} &(G(\bc-\bb^{t-i}_{\bx+\Delta}; \bo,\! \SigCB)\\
       \times G(\Delta; \bo,\! \SigXB) &\times P^{t-i}_{\bx+\Delta}(\bg|\bb^{t-i}_{\bx+\Delta})).
\end{split}
\end{equation}
Each of the data samples from the previous frames are weighted according to 
its probability of belonging to the background.
$\Sigma^B = (\SigXB,\SigCB)$ represents the covariance matrices for the 
background model.
$P^t_\bx(\bg|\bb^{t}_{\bx})$ is the probability that pixel at location 
$\bx$ in the frame $t$ is background.
$K_{\bg}$ is the appropriate normalization factor:
\begin{equation}\label{eq:bg_norm}
K_{bg} = \sum_{i\in 1:T} \sum_{\Delta\in \mNB} G(\Delta; \bo, \SigXB) \times
    P^{t-i}_{\bx+\Delta}(\bg|\bb^{t-i}_{\bx+\Delta}).
\end{equation}
For efficiency, the covariance matrices are considered to be diagonal
matrices.

\subsection{Mixing a uniform distribution component}
In cases when the background has been occluded in all the previous $T$ 
frames, there are no reliable history pixels for the background. To 
allow the system to recover from such a situation, a uniform 
color distribution is mixed into the color likelihood:
\begin{equation}
\hat{P}^t_\bx(\bc|\bg) = \gamma^\bg_\bx\times P^t_\bx(\bc|\bg;\Sigma^B) + 
(1-\gamma^\bg_\bx)\times U,
\end{equation}
where $U$ is a uniform distribution over all possible 
color values.
The mixture proportion is given by 
$\gamma^\bg_\bx = \frac{ \sum_{i\in 1:T} \sum_{\Delta\in \mNB} P^{t-i}_{\bx+\Delta}(\bg|\bb^{t-i}_{\bx+\Delta})}{\sum_{i\in 1:T} \sum_{\Delta\in \mNB} (1)}.
$
The implication of this mixture proportion is that if the history 
pixels are highly confident background pixels, then no uniform distribution
is added to the likelihood. When there is unreliable information
about background pixels in the history, a larger weight is assigned to the
uniform component.

A similar likelihood model is maintained for the foreground
process. The parameter values in our KDE implementation are
$\SigCB=\SigCF=\frac{15}{4}, \SigXB=\SigXF=\frac{5}{4}, T=5$.

\subsection{Posterior computation}
The classification results from the previous frame contain useful prior 
information about which pixels are likely to belong to the background.
The background posterior probability at each pixel in the previous 
frame is motion-compensated according to optical flow and used as the
pixelwise background prior for  the current frame.
A smoothed($7\times 7$ Gaussian filter with a standard deviation value of $1.75$) 
image of the posterior, $\tilde{P}^{t-1}_\bx(\bg)$, is used for 
the prior for the background process in the current frame.

The posterior probability of background in the current frame can now 
be computed by combining the color likelihoods, the segmentation label
likelihoods from the graphical model, and the prior:
\begin{equation}\label{eq:bayes}
P^t_\bx(\bg|\bc,l_\bx) = \frac{\hat{P}^t_\bx(\bc|\bg)\times P^t_\bx(l_\bx|\bg)\times P^t_\bx(\bg)}
{\sum_{L=\bg, \fg} \hat{P}^t_\bx(\bc|L; \Sigma^l)\times P^t_\bx(l_\bx|L)\times P^t_\bx(L)}. 
\end{equation}
The use of color likelihoods and prior information helps to recover from errors
in the FOF-based segmentation as we explain in the results.

\section{Results}\label{sec:results}
The system's performance is evaluated on two existing benchmarks. 
In addition to these benchmarks, we
also present results on a new set of videos that include several with
complex background phenomena to highlight the strengths of the
system.
The first benchmark is a motion segmentation 
data set~\cite{Brox10}, derived from the Hopkins data set~\cite{Tron07}, which 
consists of 26 moving camera videos. The data set has ground truth segmentation for
a few frames sampled throughout the video.
The second data set is the SegTrack segmentation data set~\cite{Tsai10}. 
The third data set,\footnote{The ComplexBackground data set is available for public use
at\\ \href{http://vis-www.cs.umass.edu/motionSegmentation/complexBgVideos.html}{http://vis-www.cs.umass.edu/motionSegmentation/complexBgVideos.html}.} 
which we produced ourselves, is a challenging one with complex backgrounds including
trees in a forest and large occluding objects in front of the moving 
foreground object. This data set is extremely challenging for traditional
motion segmentation algorithms.

Table \ref{tbl:fMeasures} shows the average F-measure, 
$F=\frac{2\times Rc\times Pr}{Rc+Pr}$, where $Pr$ is precision 
and $Rc$ is the recall for the background label, for each video.
We present results of FOF segmentation as well as segmentation that combines
FOF with color appearance and prior models. In general, the use of color and 
prior information helps improve the accuracy of FOF segmentation. In the 
Hopkins set, the videos that are challenging for us are the ones where
the foreground object's FOF matches the camera motion's FOF for a long 
duration (cars4), the foreground object covers a majority of the pixels in
the image (marple6, marple9), or where the foreground object is stationary 
for the first few hundred frames although the ground truth labeling considers
them to be moving because they move later on in the sequence 
(marple6, marple11).
Among the SegTrack data set, three videos (marked with *) have multiple
moving objects, but the ground truth intended for tracking analysis marks
only one primary object as the foreground, causing our system to appear
less accurate. We chose to retain the original ground truth 
labeling and report the numbers as seen.

\begin{figure*}[t]
\begin{center}
    \includegraphics[width=.8\linewidth]{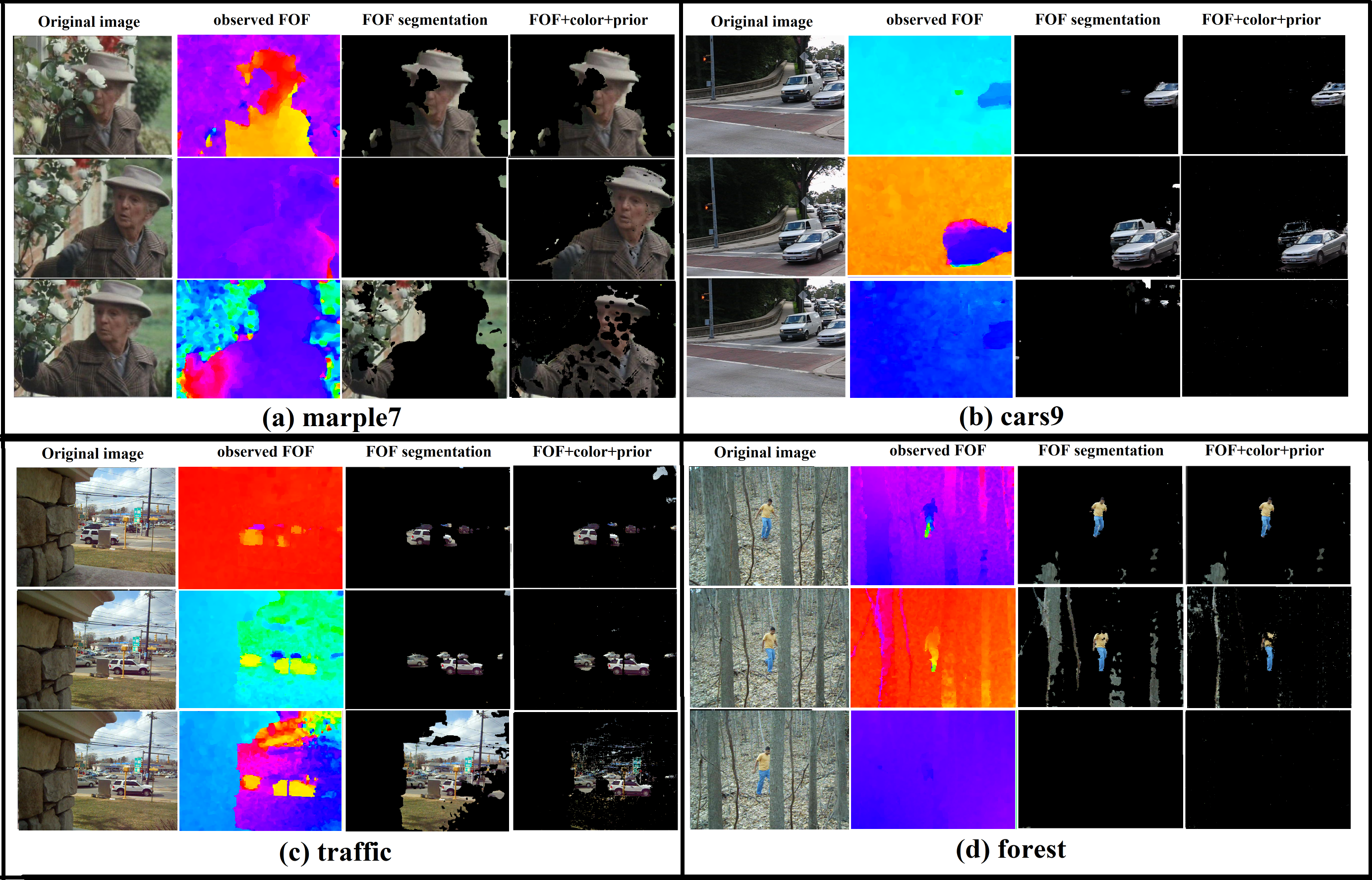}
\end{center}
   \caption{Sample results from four videos. The columns are the original
       image, the observed FOF, FOF segmentation results, and results from
       combining FOF with color and prior models, respectively.
       FOF is very accurate when the foreground objects' FOFs are easily
       distinguishable from the camera motion's FOF. When the observed
       FOF cannot distinguish between the foreground and the background, 
       FOF segmentation is not accurate. Color and prior information can
       help in these cases (row 2 in (a)). If the foreground object
       is not obvious from the FOF for a long duration, the
       color and prior too are unable to help recover them after
       some time (row 3 in (b) and (d)).
       In the new videos(c and d), camera rotation is a challenge (row 3 in (c) and row 2 in (d)). 
       Occasionally, the largest detected segment is the foreground object,
       which gets labeled as background (row 3 in (c)). 
       Using a prior helps reduce this error as well as errors due to 
       rotation. 
   }
\label{fig:sequenceResults}
\end{figure*}

Finally, in our new ComplexBackground videos taken with a hand-held camera, 
rotation is a big challenge. 
In videos where there is rotation in many frames 
(forest, drive, store), FOF segmentation is less accurate. Using color 
information helps in many of these videos. 
The forest video has the additional
challenge that the foreground object moves very slowly in many frames.
Despite these challenges in the complex background videos, 
our system performs segmentation with reasonable accuracy 
across all three data sets.
Figure \ref{fig:sequenceResults} shows a few sample segmentation results from
four videos.

The most relevant papers for foreground-background classification 
are Kwak~\etal~\cite{Kwak11}, and Elqursh and Elgammal~\cite{Elqursh12}. 
Other papers that use the Hopkins data ~\cite{Tron07,Brox10,Ochs12} 
report sparse trajectory classification results for each frame which 
are not directly comparable to foreground-background classification 
accuracy measures. 

Elqursh and Elgammal perform a spectral clustering of trajectories 
and obtain a dense labeling of pixels. However, segmentation of each frame
is performed by considering trajectory information 
from the current frame as well as four future frames.
FOF segmentation is a frame-to-frame segmentation method and hence solving a 
different problem with the aim of achieving real-time processing of frames.

Kwak~\etal report results on 3 of the 26 videos in the Hopkins data set,
where they use a special initialization procedure to segment the object 
of interest in the first frame. For the \emph{Cars1}, \emph{People1}, 
and \emph{People2} videos, they report average F-measure values of .88,
.94, and .87, respectively. Our method which makes no assumptions about the 
first frame and does not require an initialization step is not as accurate 
on the first two videos. In particular, as shown in the \emph{Cars1} video
in Figure~\ref{fig:segCompare} (last column), a heavy penalty is paid when 
our bootstrapped system fails to detect the object in the first frame. 


\begin{table}\label{tbl:fMeasures}
\small
\begin{center}
\begin{tabular}{|l|c|c|}
\hline
Videoname & FOF only & FOF+color+prior\\ 
\hline\hline
Cars1 & 47.81 & 50.84\\
Cars2 & 46.37 & 56.60\\
Cars3 & 67.18 & 73.57\\
Cars4 & 38.51 & 47.96\\
Cars5 & 64.85 & 70.94\\
Cars6 & 78.09 & 84.34\\
Cars7 & 37.63 & 42.92\\
Cars8 & 87.13 & 87.61\\
Cars9 & 68.99 & 66.38\\
Cars10 & 53.98 & 50.84\\
People1 & 56.76 & 69.53\\
People2 & 85.35 & 88.40\\
Tennis & 61.63 & 67.59\\
Marple1 & 65.65 & 88.25\\
Marple2 & 49.68 & 60.88\\
Marple3 & 67.83 & 70.71\\
Marple4 & 61.33 & 69.01\\
Marple5 & 50.05 & 45.15\\
Marple6 & 26.95 & 23.95\\
Marple7 & 51.57 & 67.13\\
Marple8 & 68.89 & 80.32\\
Marple9 & 40.53 & 36.36\\
Marple10 & 57.19 & 58.72\\
Marple11 & 37.33 & 41.41\\
Marple12 & 65.83 & 70.01\\
Marple13 & 67.09 & 80.96\\
\hline
birdfall2 & 68.68 & 75.69\\
girl  & 75.73 & 81.95\\
parachute  & 51.49 & 54.36\\
cheetah*  & 12.68 & 22.31\\
penguin*    & 14.74 & 20.71\\
monkeydog*    & 10.79 & 18.62\\
\hline
drive     & 30.13 & 61.80\\
forest    & 19.48 & 31.44\\
parking   & 43.47 & 73.19\\
store     & 28.46 & 70.74\\
traffic   & 66.08 & 71.24\\
\hline
\end{tabular}
\end{center}
\caption{Results. F-measure value for all videos in the three data sets}
\end{table}

\section{Discussion}\label{sec:discussion}
We have presented a system for motion segmentation by using 
optical flow orientations.
The use of optical flow orientations avoids the over-segmentation
of the scene into depth-dependent entities.
The system is able to automatically determine the number of foreground 
motions.
We have shown promising results on a wide range of videos including some 
with complex backgrounds.
The main drawback of our system is that it models only 
translation and is prone to
error when the camera rotates. Explicitly modeling the camera rotation
could help handle such errors.
Incorporating magnitude information can help improve
the model, especially in cases where a tracked foreground object suddenly 
disappears in the FOF observations.

\section{Acknowledgements}\label{sec:ack}
Thanks to Marwan Mattar for helpful discussions.
This work was supported in part by the National Science Foundation under CAREER award
IIS-0546666.

{\small
\bibliographystyle{ieee}
\bibliography{motionSegmentation}

\begin{thebibliography}{10}\itemsep=-1pt

\bibitem{Brox10}
T.~Brox and J.~Malik.
\newblock Object segmentation by long term analysis of point trajectories.
\newblock In {\em ECCV}, 2010.

\bibitem{Butler12}
D.~J. Butler, J.~Wulff, G.~B. Stanley, and M.~J. Black.
\newblock A naturalistic open source movie for optical flow evaluation.
\newblock In {\em ECCV}, 2012.

\bibitem{Chockalingam09}
P.~Chockalingam, S.~N. Pradeep, and S.~Birchfield.
\newblock Adaptive fragments-based tracking of non-rigid objects using level
  sets.
\newblock In {\em ICCV}, 2009.

\bibitem{Elgammal00}
A.~M. Elgammal, D.~Harwood, and L.~S. Davis.
\newblock Non-parametric model for background subtraction.
\newblock In {\em ECCV}, 2000.

\bibitem{Elqursh12}
A.~Elqursh and A.~M. Elgammal.
\newblock Online moving camera background subtraction.
\newblock In {\em ECCV}, 2012.

\bibitem{Hayman03}
E.~Hayman and J.-O. Eklundh.
\newblock Statistical background subtraction for a mobile observer.
\newblock In {\em ICCV}, 2003.

\bibitem{Irani94}
M.~Irani, B.~Rousso, and S.~Peleg.
\newblock Computing occluding and transparent motions.
\newblock {\em IJCV}, 12:5--16, 1994.

\bibitem{Irani94CVPR}
M.~Irani, B.~Rousso, and S.~Peleg.
\newblock {Recovery of ego-motion using image stabilization}.
\newblock In {\em CVPR}, 1994.

\bibitem{Kwak11}
S.~Kwak, T.~Lim, W.~Nam, B.~Han, and J.~H. Han.
\newblock Generalized background subtraction based on hybrid inference by
  belief propagation and {B}ayesian filtering.
\newblock In {\em ICCV}, 2011.

\bibitem{Lee11}
Y.~J. Lee, J.~Kim, and K.~Grauman.
\newblock Key-segments for video object segmentation.
\newblock In {\em ICCV}, 2011.

\bibitem{Narayana12BMVC}
M.~Narayana, A.~Hanson, and E.~Learned-Miller.
\newblock Improvements in joint domain-range modeling for background
  subtraction.
\newblock In {\em BMVC}, 2012.

\bibitem{Neal00}
R.~M. Neal.
\newblock Markov chain sampling methods for {D}irichlet process mixture models.
\newblock {\em Journal of Computational and Graphical Statistics},
  9(2):249--265, 2000.

\bibitem{Ochs12}
P.~Ochs and T.~Brox.
\newblock Higher order motion models and spectral clustering.
\newblock In {\em CVPR}, 2012.

\bibitem{Ren03}
Y.~Ren, C.-S. Chua, and Y.-K. Ho.
\newblock Statistical background modeling for non-stationary camera.
\newblock {\em Pattern Recognition Letters}, 24(1-3):183--196, Jan. 2003.

\bibitem{Sheikh09}
Y.~Sheikh, O.~Javed, and T.~Kanade.
\newblock Background subtraction for freely moving cameras.
\newblock In {\em ICCV}, 2009.

\bibitem{Sheikh05}
Y.~Sheikh and M.~Shah.
\newblock Bayesian modeling of dynamic scenes for object detection.
\newblock {\em PAMI}, 27, 2005.

\bibitem{Shi98}
J.~Shi and J.~Malik.
\newblock Motion segmentation and tracking using normalized cuts.
\newblock In {\em ICCV}, 1998.

\bibitem{Stauffer99}
C.~Stauffer and W.~E.~L. Grimson.
\newblock {Adaptive background mixture models for real-time tracking}.
\newblock In {\em CVPR}, 1999.

\bibitem{Sun10}
D.~Sun, S.~Roth, and M.~J. Black.
\newblock Secrets of optical flow estimation and their principles.
\newblock In {\em CVPR}, 2010.

\bibitem{Tron07}
R.~Tron and R.~Vidal.
\newblock A benchmark for the comparison of 3-d motion segmentation algorithms.
\newblock In {\em CVPR}, 2007.

\bibitem{Tsai10}
D.~Tsai, M.~Flagg, and J.~M. Rehg.
\newblock Motion coherent tracking with multi-label {MRF} optimization.
\newblock {\em BMVC}, 2010.

\end{thebibliography}
}

\end{document}